\documentclass{article}

\usepackage{microtype}
\usepackage{graphicx}
\usepackage{subfigure}
\usepackage{booktabs} 

\usepackage{times}
\usepackage{latexsym}
\usepackage{float}
\usepackage{makecell}
\usepackage[T1]{fontenc}

\usepackage[utf8]{inputenc}

\usepackage{microtype}

\usepackage{inconsolata}

\usepackage{hyperref}
\usepackage{multirow}

\usepackage[accepted]{icml2025}

\usepackage{amsmath}
\usepackage{amssymb}
\usepackage{mathtools}
\usepackage{amsthm}
\usepackage[capitalize,noabbrev]{cleveref}

\theoremstyle{plain}

\theoremstyle{definition}

\theoremstyle{remark}

\newcommand{\ie}{\textit{i.e.}}
\newcommand{\eg}{\textit{e.g.}}

\newcommand{\ours}{{HALO}}

\usepackage[textsize=tiny]{todonotes}
\usepackage[most]{tcolorbox}
\definecolor{midnightgreen}{rgb}{0.0, 0.29, 0.33}
\definecolor{darkpurple}{rgb}{0.31, 0.18, 0.43}

\icmltitlerunning{Harness Local Rewards for Global Benefits: Effective Text-to-Video Generation Alignment with Patch-level Reward Models}

\begin{document}

\twocolumn[
\icmltitle{Harness Local Rewards for Global Benefits: Effective Text-to-Video Generation Alignment with Patch-level Reward Models}



\icmlsetsymbol{equal}{*}

\begin{icmlauthorlist}
\icmlauthor{Shuting Wang}{a1,a2,equal}
\icmlauthor{Haihong Tang}{a3}
\icmlauthor{Zhicheng Dou}{a2}
\icmlauthor{Chenyan Xiong}{a1}
\end{icmlauthorlist}

\icmlaffiliation{a1}{School of Computer Science, Carnegie Mellon University}
\icmlaffiliation{a2}{Gaoling School of Artificial Intelligence, Renmin University of China}
\icmlaffiliation{a3}{Serendipity One Inc.}
\icmlcorrespondingauthor{Shuting Wang}{wangshuting@ruc.edu.cn}
\icmlcorrespondingauthor{Chenyan Xiong}{cx@cs.cmu.edu}

\icmlkeywords{Machine Learning, ICML}

\vskip 0.3in
]



\printAffiliationsAndNotice{\icmlEqualContribution} 

\begin{abstract}
The emergence of diffusion models (DMs) has significantly improved the quality of text-to-video generation models (VGMs). However, current VGM optimization primarily emphasizes the global quality of videos, overlooking localized errors, which leads to suboptimal generation capabilities. 
To address this issue, we propose a post-training strategy for VGMs, \ours{}, which explicitly incorporates local feedback from a patch reward model, providing detailed and comprehensive training signals with the video reward model for advanced VGM optimization. 
To develop an effective patch reward model, we distill GPT-4o to continuously train our video reward model, which enhances training efficiency and ensures consistency between video and patch reward distributions. Furthermore, to harmoniously integrate patch rewards into VGM optimization, we introduce a granular DPO (Gran-DPO) algorithm for DMs, allowing collaborative use of both patch and video rewards during the optimization process.
Experimental results indicate that our patch reward model aligns well with human annotations and \ours{} substantially outperforms the baselines across two evaluation methods. Further experiments quantitatively prove the existence of patch defects, and our proposed method could effectively alleviate this issue.
\end{abstract}

\section{Introduction}\label{sec:intro}
\begin{figure}
    \centering
    \includegraphics[width=1\linewidth]{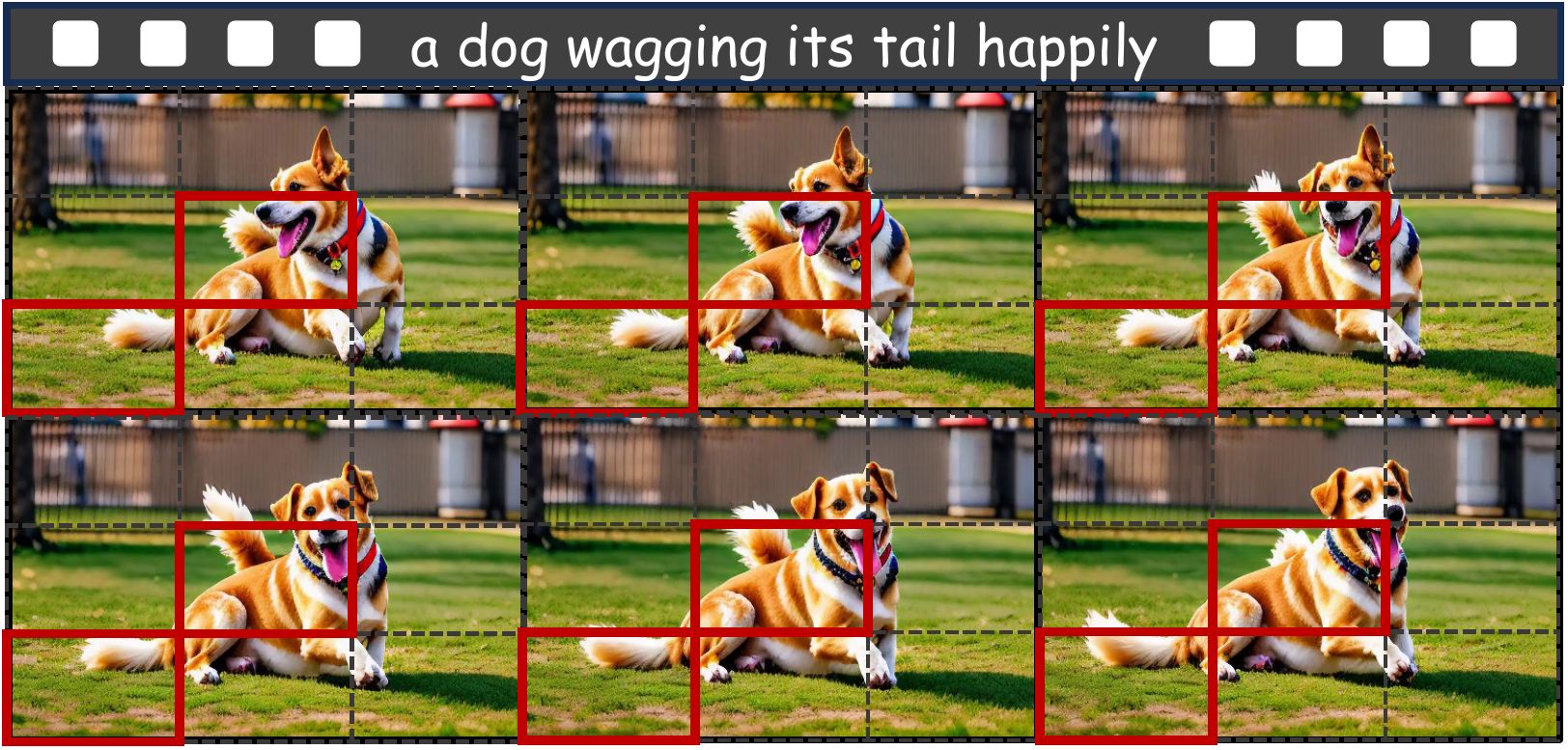}
    \caption{A generated video with local flaws (red boxed). 
    }
    \vspace{-0.5cm}
    \label{fig:patch-defect-case}
\end{figure}

The advent of diffusion models has significantly elevated the quality of AI-generated videos, enabling the creation of high-resolution and visually appealing AI videos. 
This advancement holds great promise for various industrial applications, such as video editing and artistic creation~\cite{Pictory,runwayml,synthesia}.

Typically, the pre-training of VGMs relies on extensive text-to-video datasets~\cite{webvid-1m}, leading to strong foundational generation abilities. However, the varied quality of these datasets still limits the potential of VGMs. 
Inspired by the alignment of large language models (LLMs)~\cite{rlhf}, recent studies~\cite{t2v-turbo} introduced reward models to assess the video generation quality. This approach facilitates the alignment of VGMs during the post-training stages and improves the overall generation quality. 

Although current VGMs can generally yield visually appealing videos, they still exhibit spatially localized errors in certain video patches. 
Unfortunately, these localized errors manifest significant hallucinations that diverge from reality, resulting in poor user experience and awkward video generation. 
For example, as illustrated in Figure~\ref{fig:patch-defect-case}, when given the input prompt ``a dog wagging its tail happily'', an advanced VGM (T2V-Turbo-v1~\cite{t2v-turbo}) generates a video that \textit{appears commendable at first glance}. However, in the video patches indicated by red boxes, there are \textit{two tails} for the dog, which contradicts real-world expectations. 

In this paper, we propose a post-training framework, \ours{}, that explicitly considers patch-level feedback for generated videos to enhance the alignment of VGMs. 
Specifically, our framework \textbf{HA}rnesses \textbf{LO}cal feedback from a specialized patch reward model to identify localized and fine-grained defects in generated videos. 
We integrate it with a video reward model that assesses the global quality of generated videos, thereby ensuring that models do not prioritize local details at the expense of the overall quality of videos.

To implement our approach, we first distill GPT-4o~\cite{GPT-4} to continuously optimize the video reward model, resulting in our patch reward model. This strategy leverages the video reward model's foundational ability to evaluate the video quality, thereby reducing the burden of data labeling and training resource allocation. Furthermore, originating from the same model enables patch and reward models to generate consistent reward distributions and scales, minimizing potential conflicts and providing uniform guidance during the optimization of VGMs. 
Next, to integrate patch rewards into VGM optimization, we develop a novel Gran-DPO algorithm. This algorithm first tailors a patch DPO from the classic Diffusion-DPO~\cite{DMDPO}, which serves as the video DPO. Then, the patch and video DPO losses are weightily combined to achieve a harmonious optimization of both local and global generation quality.

We conduct experiments using the VBench~\cite{vbench} prompts and evaluate models by VBench and VideoScore~\cite{VideoScore}. Experimental results confirm that \ours{} significantly outperforms all baselines. Further analyses reveal several key findings: \textit{First}, the patch reward model demonstrates positive correlations with human annotations, validating the efficacy of our optimization signals. \textit{Next}, patch reward preferences exhibit certain differences from video reward preferences, suggesting that patch rewards could introduce unique values beyond video rewards. 
\textit{Finally}, \ours{} effectively improves patch reward values, indicating its abilities in dealing with local defects.

The main contributions of our study are threefold:

(1) We introduce a patch reward model to produce localized feedback for generated videos, facilitating a fine-grained VGM optimization.

(2) We develop a Gran-DPO algorithm to leverage local and global rewards collaboratively, hence optimizing VGMs in a harmonious manner.

(3) Experiments prove the consistency of our patch reward model with human annotations and the capability of our method to address the problem of local defects.
\section{Related Work}\label{{sec:related}}
\paragraph{Diffusion-based VGMs.}
Recently, with the superiority of diffusion models in image generation tasks~\cite{DM1,DM2,DDIM,SDXL,LatentDM}, many studies also proposed DM-based text-to-video generation models~\cite{VDM,LAVIE,Text2Video-Zero,VC2,CogVideoX,ModelScope}. 
The structures of DMs typically have two main branches. One is UNet-based models~\cite{UNet,VC2,ModelScope}, such as VideoCrafter1~\cite{VC1}, which used 3D-UNet~\cite{3DUNet} to effectively model temporal information. Another is transformer-based models~\cite{DiT,EasyAnimate-DiT,CogVideo,CogVideoX}, such as CogVideo~\cite{CogVideo,CogVideoX}, which stacked expert transformer-blocks to implement diffusion denoising. 

Due to the limited amount and quality of text-to-video datasets~\cite{webvid-1m,OpenVid-1M,UCF101} compared to text-to-image datasets~\cite{LAION-5B}, some works used pre-trained image generation models to build VGMs~\cite{show-1,AlignLatent,Text2Video-Zero}, or combined image and video datasets to optimize VGMs~\cite{VC1,VC2}, leading to better foundational ability of VGMs.

\paragraph{Alignment of generative AI models.}
Fine-tuning pre-trained generative models with reward models has proven effective in enhancing their overall generation quality. 
It has been successfully applied in various fields \eg, text generation~\cite{LMFI,GPT-3.5,GPT-4,richrag} and visual generation~\cite{AlignDM,DRaFT,DPOK,DMDPO,RLDMDiverse,t2v-turbo,t2v-turbo-v2,InstructVideo}. 

Recent studies in text generation~\cite{TLCR,Step-DPO} have gone beyond rewarding entire generated responses. They introduced local feedback at the step or token level to provide fine-grained signals, enhancing LLMs' reasoning and generation capabilities. 
For text-to-video generation, advanced studies also leveraged reward models to evaluate the global quality of generated videos. For example, some studies~\cite{DRaFT,DDPO,DPOK,t2v-turbo,InstructVideo} used HPSv-2~\cite{HPSv2} or PickScore~\cite{PickScore} to measure the alignment between text and images. Recently, VideoScore~\cite{VideoScore} proposed to evaluate video quality by fine-tuning visual LLMs with human-annotated labels. 
Despite these advancements, there is still room for exploring the role of local feedback in videos to enhance the abilities of VGMs. In this study, we propose a new post-training framework for VGMs that incorporates patch rewards, enabling a better quality of the generated videos.
\section{\ours{} Method}\label{sec:method}

\begin{figure*}[ht]
    \centering
    \includegraphics[width=1\linewidth]{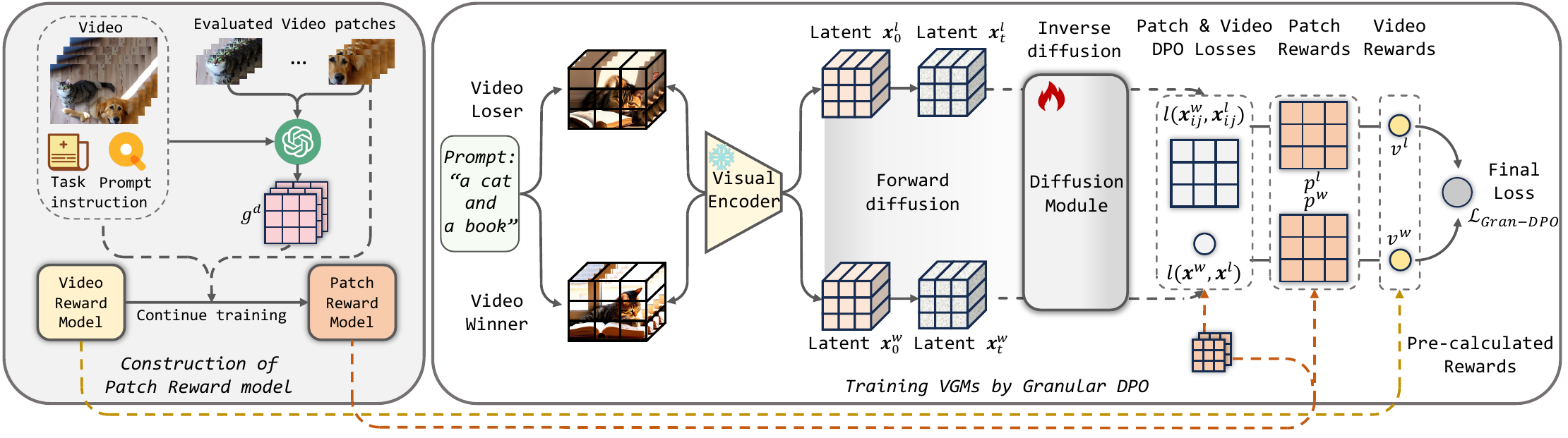}
    \caption{The framework of our proposed method, \ours{}.
    }
    \label{fig:model}
\end{figure*}

In this section, we detail the techniques of our proposed method, \ours{}. 
Firstly, we provide essential definitions and functions of diffusion models and their DPO algorithms in Section~\ref{sec:preliminary}, serving as the background of our study.
Next, we introduce the construction of our reward model in Section~\ref{sec:rw}. 
Following that, Section~\ref{sec:dpo} illustrates the details of our Gran-DPO algorithm for diffusion models. Finally, the acquisition of training prompts is presented in Section~\ref{sec:prompt}. The framework of \ours{} is visualized in Figure~\ref{fig:model}.

\subsection{Preliminaries}\label{sec:preliminary}

\paragraph{Text-to-video diffusion models.}
Generally, a text-to-video generation model comprises three key modules: (1) A text encoder $\mathcal{C}$, which embeds the text prompt into a latent representation $\mathbf{c}$, serving as the conditional embedding for the diffusion process. (2) A video autoencoder $\mathcal{E}$-$\mathcal{D}$. It encodes a video $\mathbf{v}$ into (or decodes from) the latent space by 
$\mathbf{x}=\mathcal{E}(\mathbf{v}), \mathbf{v}=\mathcal{D}(\mathbf{x})$, $\mathbf{x}\in\mathbb{R}^{f\times h\times w\times c},\mathbf{v}\in\mathbb{R}^{F\times H\times W\times 3}$. $f$,$h$, and $w$ are the dimensions compressed from $F$,$H$, and $W$, representing the frame number, height, and width of videos, respectively. $c$ is the output channel number of $\mathcal{E}$. 
(3) A denoising module $\epsilon_\theta$, which conducts the forward and reverse diffusion processes to implement the training and inference of diffusion models. 

For training the diffusion module, the input text and video are firstly embedded into latent representations $(\mathbf{c}, \mathbf{x})$ ($\mathbf{x}$ also termed as $\mathbf{x}_0$). By a predefined scheduler $\{\alpha_t\}_{t=1}^T$, the forward diffusion adds random noise to $\mathbf{x}_0$ to build $\mathbf{x}_t$: 
\begin{equation}
     \mathbf{x}_t = \sqrt{\overline{\alpha}_t}\mathbf{x}_0 + \sqrt{1-\overline{\alpha}_t}\epsilon, \epsilon\sim\mathcal{N}(\mathbf{0},\mathbf{I}),t\sim\mathcal{U}(1,T).
\end{equation}
Then, the denoising module performs the reverse diffusion to model the noise at the $t$-th step conditioned on the text embedding. The loss function~\cite{DM2} is:
\begin{equation}
    \mathcal{L}(\theta) = \mathbb{E}_{\mathbf{x}, \mathbf{c}, \epsilon, t}\left[\|\epsilon - \epsilon_\theta(\mathbf{x}_t, \mathbf{c}, t) \|^2_2\right].
\end{equation}
The inference of the diffusion process starts by normally sampling the initial latent $\mathbf{x}_T\sim\mathcal{N}(\mathbf{0},\mathbf{I})$. Then it generates the video latent $\mathbf{x}_0$ via a stepwise reverse diffusion process: 
\begin{align}
    \mathbf{x}_{t-1} &= \frac{1}{\sqrt{\alpha_t}}(\mathbf{x}_t-\frac{1-\alpha_t}{\sqrt{1-\overline{\alpha_t}}}\epsilon_\theta(\mathbf{x}_t, t))+\sigma_t\mathbf{z}, \nonumber\\
    \mathbf{z}&\sim\mathcal{N}(\mathbf{0},\mathbf{1}), \;\;\text{if}\;\; t>1, \;\;\text{else}\;\; \mathbf{z}=\mathbf{0}.
\end{align}
To accelerate the inference, DDIM sampling~\cite{DDIM} is widely used to generate videos by fewer sampling steps than training.

\paragraph{DPO for diffusion models.}
The DPO algorithm has been widely adopted in the fine-tuning stage of LLMs, benefiting from its easy implementation and stable optimization. Recently, \citet{DPODM} also tailored DPO to diffusion models. The derived loss function is presented as follows:

\begin{align}
\small
    \mathcal{L}_{\text{DM-DPO}}(\theta) = -\mathbb{E}_{(\mathbf{x}^w,\mathbf{x}^l),t,(\mathbf{x}_t^w,\mathbf{x}_t^l)}\left[
    \log \sigma \left(
    -\beta T\left( \nonumber \right.\right.\right.\\ 
    \|\epsilon^w-\epsilon_\theta(\mathbf{x}^w_t ,\mathbf{c},t) \|^2_2-
    \|\epsilon^w-\epsilon_{\text{ref}}(\mathbf{x}^w_t,\mathbf{c},t) \|^2_2-\nonumber \\ \left.\left.\left.
    \left(
    \|\epsilon^l-\epsilon_\theta(\mathbf{x}^l_t ,\mathbf{c},t) \|^2_2-
    \|\epsilon^l-\epsilon_{\text{ref}}(\mathbf{x}^l_t,\mathbf{c},t) \|^2_2
    \right)
    \right) 
    \right) 
    \right].
\label{eq:diffusion_dpo}
\end{align}

$\mathbf{x}^w$ and $\mathbf{x}^l$ are the winner and the loser within a video pair, $\epsilon_{\text{ref}}$ is the reference model, $\sigma()$ denotes the sigmoid function, and $\beta$ is a hyperparameter to control regularization. The objective of this optimization is to enhance the diffusion module to optimize more on the winner $\mathbf{x}^w$ than the loser $\mathbf{x}^l$ while ensuring the optimized policy model $\epsilon_\theta$ remains close to the reference model $\epsilon_{\text{ref}}$. Following~\cite{DM2}, we simplify optimization by removing the weighting function.

\subsection{Construction of Patch Reward Model}\label{sec:rw}
We use VideoScore~\cite{VideoScore} as our video reward model, which evaluates the video quality by a fine-tuned visual LLM, Mantis-Idefics2-8B~\cite{mantis}. The fine-tuning process relies on human-annotated scores for AI-generated videos. VideoScore defines five evaluation dimensions, such as visual quality, text-to-video alignment, etc. For an input text-video pair, it predicts five scores on a scale of 1 to 4. We treat the average of five evaluation scores to derive the final video reward score, $v$:
\begin{equation}
    v = \frac{1}{|\mathcal{D}^{VS}|}\sum\nolimits_{d\in\mathcal{D}^{VS}}v^d.
    \label{eq:rw_avg}
\end{equation}
$\mathcal{D}^{VS}$ is the set of evaluation dimensions and $v^d$ denotes the reward score of the video for the specific dimension $d$.

Considering the lack of publicly available patch reward models, we propose fine-tuning our video reward model to create the patch reward model. This operation allows us to preserve similar distributions and scales between patch and video reward scores, while also simplifying the training task by leveraging the foundational evaluation capabilities of the video reward model. To mitigate the time and economic costs associated with human annotation, we employ GPT-4o~\cite{GPT-4} to label the quality of video patches and then distill it to build our patch reward model.  

\paragraph{GPT-4o labeling.} We divide a video into a $h_n\times w_n$ grid of video patch alongside height and width, where $h_n$ and $w_n$ are the number of divisions in height and width. Note that we do not segment videos along the time axis, as current AI-generated videos are relatively short (usually only a few seconds), resulting in a low information density across timestamps. We leave the task of temporal segmentation for future work. For segmented video patches, we define their indexes by referring to the upper left corner as $(0,0)$.

To ensure complete semantics during the evaluation, we also inject the entire video into the input, thereby contextualizing the evaluated video patch. Ultimately, we treat the task instruction, the text prompt, the video, a certain video patch, and its index as input for evaluation. Following VideoScore, we prompt GPT-4o to generate five-dimensional evaluation scores, which serve as training labels. The instruction for GPT-4o is presented in Appendix~\ref{app:instructions}.\footnote{We instruct GPT-4o to generate evaluation scores from 0 to 10 to mitigate the issue of identical scores. Then, these scores are normalized to a scale of 1 to 4 to align with VideoScore.} 

\paragraph{Distill GPT-4o to build the patch RM.} With the labeled data from GPT-4o, we fine-tune the video reward model by regression loss following VideoScore, as expressed below,
\begin{equation}\small
    \mathcal{L}_{\text{reg}}(\mathbf{v}) =\frac{1}{h_nw_n|\mathcal{D}^{VS}|}\sum_{i=1}^{h_n}\sum_{j=1}^{w_n}\sum_{d\in\mathcal{D}^{VS}} \| p^{d}_{ij} - g^{d}_{ij} \|_2^2.
\end{equation}
$p^{d}_{ij}$ denotes the reward score predicted by the patch reward model for the video patch in the index of $(i,j)$, and $g^{d}_{ij}$ is the GPT-4o-labeled score. The final patch reward $p_{ij}$ is also the average score of all dimensions similar to Eq.~(\ref{eq:rw_avg}).

\subsection{Granular DPO for VGMs}\label{sec:dpo}

\paragraph{Building pairwise training data for Gran-DPO.} 
Given the set of training prompts (the creation is introduced in Section~\ref{sec:prompt}), we first input every prompt into our base text-to-video generation models and sample several generated videos, thereby constructing the training text-to-video instances. Next, we employ our two specialized reward models to compute patch and video rewards for each instance. Finally, we develop a data builder to create the pairwise training instances for the Gran-DPO algorithm. Specifically, for a text prompt and its generated videos, we perform a pairwise comparison of their rewards and retain data pairs that satisfy one of the following criteria: 
(1) The video reward margin between two videos exceeds the statistical median of all pairwise video reward margins $m_V$.
(2) One of the patch reward margins surpasses the statistical median of all pairwise patch reward margins $m_P$.

\paragraph{Gran-DPO algorithm.} 

To collaboratively align VGMs from both local and global perspectives, we adapt the Diffusion-DPO algorithm to create the Gran-DPO algorithm. This approach integrates patch DPO and video DPO losses to formulate the final optimization object.

Specifically, given a pair of text-to-video instances, their initial and $t$-th latents are $\mathbf{x}^w, \mathbf{x}^l, \mathbf{x}^w_t, \mathbf{x}^l_t\in\mathbb{R}^{f\times h\times w\times c}$. The ``winner'' and ``loser'' are determined by comparing their video rewards. Following Eq.~(\ref{eq:diffusion_dpo}), the video DPO loss for this pair can be represented as below:
\begin{align}\small
     l(\mathbf{x}^w,\mathbf{x}^l) &= -\log \sigma(
    -\beta T( \nonumber\\
    \|\epsilon^w  -\epsilon_\theta&(\mathbf{x}^w_t,\mathbf{c},t) \|^2_2-
    \|\epsilon^w-\epsilon_{\text{ref}}(\mathbf{x}^w_t,\mathbf{c},t) \|^2_2-\nonumber\\
    (
    \|\epsilon^l\; -\epsilon_\theta&(\mathbf{x}^l_t\;,\mathbf{c},t) \|^2_2-
    \|\epsilon^l\;-\epsilon_{\text{ref}}(\mathbf{x}^l_t\;,\mathbf{c},t) \|^2_2
    )
    ) 
    ). 
    \label{eq:dpo-pair}
\end{align}
To calculate patch DPO loss, we first split latents along the height and width into a $h_n\times w_n$ grid of patch latents. The shape of a patch latent, $\mathbf{x}_{ij}$, is $f\times\lfloor \frac{h}{h_n}\rfloor\times\lfloor \frac{w}{w_n}\rfloor\times c$.\footnote{The height (width) of patches in the last row(column) is $h-(h_n-1)\lfloor\frac{h}{h_n}\rfloor$ ($w-(w_n-1)\lfloor\frac{w}{w_n}\rfloor$).} Therefore, for a patch pair within the given video pair, we construct their patch DPO loss by adapting Eq.~(\ref{eq:dpo-pair}):
\begin{align}\small
    l(\mathbf{x}^{w}_{ij},\mathbf{x}^{l}_{ij}) &= -\log \sigma(
    -\beta T \cdot \mathbb{F}(p^{w}_{ij},p^{l}_{ij})( \nonumber\\
    \|\epsilon^{w}_{ij}  -\epsilon_\theta&(\mathbf{x}^w_t,\mathbf{c},t)_{ij} \|^2_2-
    \|\epsilon^{w}_{ij}-\epsilon_{\text{ref}}(\mathbf{x}^w_t,\mathbf{c},t)_{ij} \|^2_2-\nonumber\\
    (
    \|\epsilon^l_{ij}\; -\epsilon_\theta&(\mathbf{x}^l_t\;,\mathbf{c},t)_{ij} \|^2_2-
    \|\epsilon^l_{ij}\;-\epsilon_{\text{ref}}(\mathbf{x}^l_t\;,\mathbf{c},t)_{ij} \|^2_2
    )
    ) 
    ), \nonumber\\
    \mathbb{F}(p^{w}_{ij},p^{l}_{ij}) &= \left\{
        \begin{aligned}
            1, & \;\;\text{if}\;\; p^{w}_{ij} > p^{l}_{ij},\\
            0, & \;\;\text{if}\;\; p^{w}_{ij} = p^{l}_{ij},\\
           -1, & \;\;\text{if}\;\; p^{w}_{ij} < p^{l}_{ij}.
        \end{aligned}
        \right.
\end{align}
Note that the patch pair reward preference of the patch pair is not always consistent with the video pair. Thereby, we introduce a triple indication function $\mathbb{F}()$ to identify the actual winner and loser of the patch pair. 

Furthermore, considering that a larger reward margin indicates a more distinct quality discrepancy between patch (or video) pairs, we project the reward margin as the pair weight to make the VGM focus on important data pairs. Finally, the loss function of Gran-DPO is:
\begin{align}\small
    \mathcal{L}_{\text{Gran-DPO}}(\theta) &= -\mathbb{E}_{(\mathbf{x}_0^w,\mathbf{x}_0^l),t,(\mathbf{x}_t^w,\mathbf{x}_t^l)}
    [
    \omega(v^w,v^l)l(\mathbf{x}^w,\mathbf{x}^l)+ \nonumber\\ 
    &\quad\quad\sum\nolimits_{i,j}\omega(p^{w}_{ij},p^{l}_{ij})l(\mathbf{x}^{w}_{ij},\mathbf{x}^{l}_{ij})
    ], \nonumber\\
    \omega(v^{w},v^{l}) &= \max\left(\min\left(|v^{w}-v^{l}|/m_P, 1\right), 0\right).
\end{align}
Recall that $v^w$ and $v^l$ represent the video rewards of the winner and loser, which are calculated by Eq.~(\ref{eq:rw_avg}).

\subsection{Acquisition of Training Prompts}\label{sec:prompt}
\paragraph{Generation of text-to-video training pairs.}
To avoid overlapping with the evaluation text prompts and to ensure controllability over the number of training instances, following self-instruct~\cite{self-instruct}, we utilize GPT-3.5-Turbo~\cite{GPT-3.5} to generate text prompts, which serve as the training text prompts for our method.

Adhering to the VBench prompt format~\cite{vbench}, we generate our training prompts based on 16 evaluation dimensions \eg, image quality and multiple objects, which also improves the diversity of our training data. For each dimension, we first sample some VBench prompts to create demonstrations for GPT-3.5-Turbo. As a result, we regard the instruction, demonstrations, and dimension descriptions as the input for generating our training prompts. 
The instruction content is presented in Appendix~\ref{app:instructions}.

We further establish a filtering process to eliminate generated prompts that exhibit high similarity with existing ones, thereby avoiding data leakage and repetition issues:
\begin{align}
\mathcal{T}^T &= \{t\;|\;t\in\mathcal{T}^G \land S(t)=1\}, \\
S(t)&= \mathbb{I}\left(\max\left(\{\mathrm{sim}\left(t,\tilde{t}\right)\;|\;\tilde{t}\in\mathcal{T}^E\cup\mathcal{T}^G_{<t} \}\right) < \tau\right), \nonumber
\end{align}
where $t$ denotes a generated prompt, $\mathcal{T}^T$, $\mathcal{T}^G$, and $\mathcal{T}^E$ denote the training, generated, and evaluation prompt sets separately. $\tau$ is a hyperparameter. These generated prompts are then input to VGMs to build text-to-video instances, supporting the optimization of VGMs.

\section{Experiment Settings}\label{sec:experimetn}

\begin{table*}[h]
    \centering
    \caption{Overall experimental results under the evaluation of VBench and VideoScore. The best results are indicated in bold. Results that are better than the base models are highlighted with underscores.
    }
    \resizebox{0.99\linewidth}{!}{\begin{tabular}{lcccccccccccc}
        \toprule
        \multirow{2}{*}{Models} & \multicolumn{6}{c}{VBench} & \multicolumn{6}{c}{VideoScore} \\
        \cmidrule(lr){2-7} \cmidrule(lr){8-13}
        & \makecell{Imaging\\Quality} & \makecell{Multiple\\Objects} & \makecell{Human\\Action} & \makecell{Spatial\\Relationship} & \makecell{Scene} & Avg. & \makecell{Visual\\Quality} & \makecell{Temporal\\Consistency} & \makecell{Dynamic\\Degree} & \makecell{Text-to-video\\Alignment} & \makecell{Factual\\Consistency} & Avg. \\
        \midrule
        Lavie & 61.90 & 33.32 & 96.80 & 34.09 & 52.69 & 55.76 & 2.5655 & 2.2523 & 2.8385 & 2.4456 & 2.1970 & 2.4598 \\
        ModelScope & 58.57 & 38.98 & 92.40 & 33.68 & 39.26 & 52.58 & 2.0777 & 2.1142 & 2.7028 & 2.0670 & 1.8980 & 2.1719 \\
        VideoCrafter2 & 67.22 & 40.66 & 95.00 & 55.29 & 25.13 & 56.66 & 2.6238 & 2.3486 & 2.7142 & 2.4670 & 2.2096 & 2.4726 \\
         \midrule
        T2V-Turbo-v1 & 72.09 & 51.30 & 95.00 & 40.35 &  \textbf{56.35} & 63.02 & 2.6388 & 2.4419 & 2.8022 & 2.5744 & 2.2944 & 2.5503 \\
        \makecell{\quad + HALO} & 72.07 & \underline{54.97} & \underline{95.00} & \underline{41.10} & 54.72 & \underline{63.57} & \underline{2.6430} & \underline{2.4453} & \underline{2.8093} & \underline{2.5775} &  2.2880 & \underline{2.5526} \\
        \midrule
        T2V-Turbo-v2 & \textbf{72.50} & 59.16 & 96.40 & 35.58  & 53.60 & 63.45  & 2.5076 & 2.2961 & 2.8507 & 2.5132 & 2.2219 & 2.4779  \\
        \makecell{\quad + HALO} & 69.11 & \underline{67.50} & \underline{97.60} & \underline{53.19} & \underline{55.07} & \underline{68.49}& \underline{2.5508} & \underline{2.3502} & 2.8417 & \underline{2.5694} & \underline{2.2252} & \underline{2.5074} \\
        \midrule
        CogvideoX-2B       & 60.88 & 68.72 & 97.80 & 65.23 & 51.09 & 68.74 & 2.8128 & 2.6666 & 2.8571 & 2.7875 & 2.5045 & 2.7257 \\
        \makecell{\quad + HALO} & \underline{61.90} & \textbf{\underline{72.91}} & \textbf{\underline{98.00}} & \textbf{\underline{65.24}} & \underline{51.16} & \textbf{\underline{69.84}} & \textbf{\underline{2.8510}} & \textbf{\underline{2.6887}} & \textbf{\underline{2.9160}} & \textbf{\underline{2.8396}} & \textbf{\underline{2.5211}} & \textbf{\underline{2.7633}} \\
        \bottomrule
    \end{tabular}}
    \label{tab:overall}
\end{table*}

\paragraph{Evaluation prompts and methods.} 
In this section, we demonstrate our experimental results and further analyses to sufficiently validate the effectiveness of our proposed method, \ours{}. We conduct our experiments on VBench~\cite{vbench} prompts and use VBench and VideoScore~\cite{VideoScore} as our evaluation methods. VBench is a widely used benchmark to assess the capabilities of VGMs from 16 predefined dimensions. We choose five dimensions: image quality, multiple objects, human action, spatial relationship, and scene, which we focus on and have optimized space, to evaluate our method.\footnote{\href{https://github.com/Vchitect/VBench/tree/master}{https://github.com/Vchitect/VBench/tree/master}}  VideoScore~\cite{VideoScore} is a video quality evaluator fine-tuned from a visual LLM, Mantis-8B-Idefics2~\cite{mantis}. Specifically, \href{https://huggingface.co/TIGER-Lab/VideoScore-v1.1}{VideoScore-v1.1} serves as our evaluator. It assesses the video quality from five dimensions: visual quality, temporal consistency, dynamic degree, text-to-video alignment, and factual consistency.

\paragraph{Baselines.} To implement our method, we select three VGMs as our base models: CogVideoX-2B~\cite{CogVideoX}, which utilizes the transformer-based diffusion module; T2V-Turbo-v1 and T2V-Turbo-v2~\cite{t2v-turbo,t2v-turbo-v2}, which leverage video RMs and consistency models~\cite{CM} to post-train VideoCrafter-2.0~\cite{VC2}. Notably, all training videos in our study were generated by base models, rather than using real videos collected from the web, which is common practice in existing post-training of VGMs with video RMs~\cite{t2v-turbo,t2v-turbo-v2}. 
This approach allows us to explore the upper limits of the base models through themselves. Therefore, we evaluate our method as a plug-in across various base models. Additionally, we assess the performance of several popular VGMs, \ie, Lavie~\cite{LAVIE}, ModelScope~\cite{ModelScope}, and VideoCrafter-2.0, to highlight the superior capabilities of our base models. 

\paragraph{Implementation details.} For training data generation, 878 text prompts were produced from GPT-3.5-Turbo. Then, we enable our base models to sample five videos for each prompt. To build video patches, we set $h_n=3$ and $w_n=3$. We first prompt GPT-4o to evaluate video patches from 492 text-video pairs (sampled from training data). It contains 4,428 labeled video patches. For continuously training VideoScore-v1.1, the batch size is 64, the learning rate is 1e-6, and the epoch is 10. Then, we use our reward models to assess all text-video instances, hence building training pairs for Gran-DPO. For efficient and effective training, we use the LoRA technique~\cite{lora} to fine-tune base models. The LoRA rank was set by 64, the learning rate was set by 1e-4, and the training step was 8k. 
Following the setting of base models, the inference diffusion steps of T2V-Turbo-v1 and v2 are 8, and CogVideoX-2B is 50. 
Due to limited space, more details are demonstrated in Appendix~\ref{app:imp_detail}.

\begin{table}[]
    \centering
    \caption{Comparison results of ablation experiments. 
    }
    \resizebox{0.99\linewidth}{!}{
    \begin{tabular}{lcccc}
    \toprule
        \multirow{2}{*}{Models} & \multicolumn{2}{c}{VBench} & \multicolumn{2}{c}{VideoScore} \\
    \cmidrule(lr){2-3}\cmidrule(lr){4-5}
        & Average & Difference &  Average & Difference \\
    \midrule
        \ours{}        & 68.49 & - & 2.5074 & - \\
    \midrule
        \quad w/o PatchDPO & 63.25 & -5.24 & 2.4602 & -0.0472 \\ 
        \quad w/o VideoDPO & 63.27 & -5.22 & 2.4791 & -0.0283\\
        \quad w/o PairWeight  & 62.86 & -5.63 & 2.4891 & -0.0183\\
        \quad w/ PickScore  & 64.27 & -4.22 & 2.4981 & -0.0093\\
        \quad w/ HPSv2 & 64.63 & -3.86 & 2.4833 & -0.0241\\
    \bottomrule
    \end{tabular}
    }
    \label{tab:ablation}
\end{table}

\begin{table}[!h]
    \centering
    \small
    \caption{Consistency evaluation of patch reward models.
    }
    \resizebox{0.99\linewidth}{!}{
    \begin{tabular}{p{0.15\linewidth}p{0.35\linewidth}>{\centering\arraybackslash}p{0.4\linewidth}}
    \toprule
        Model A &  Model B & Spearman Correlations \\
    \midrule
        Human & GPT-4o & 0.4023 \\
        GPT-4o & Patch Reward Model & 0.6062 \\
        Human & Patch Reward Model & 0.3684 \\
    \bottomrule
    \end{tabular}
    }
    \label{tab:eval_patch_rw}
\end{table}
\section{Experimental Results}
In this section, we demonstrate our experimental results and further analyses to validate the effectiveness of \ours{}.
\subsection{Overall Results}\label{sec:overall}
\begin{figure*}[h]
    \centering
    \includegraphics[width=1\textwidth]{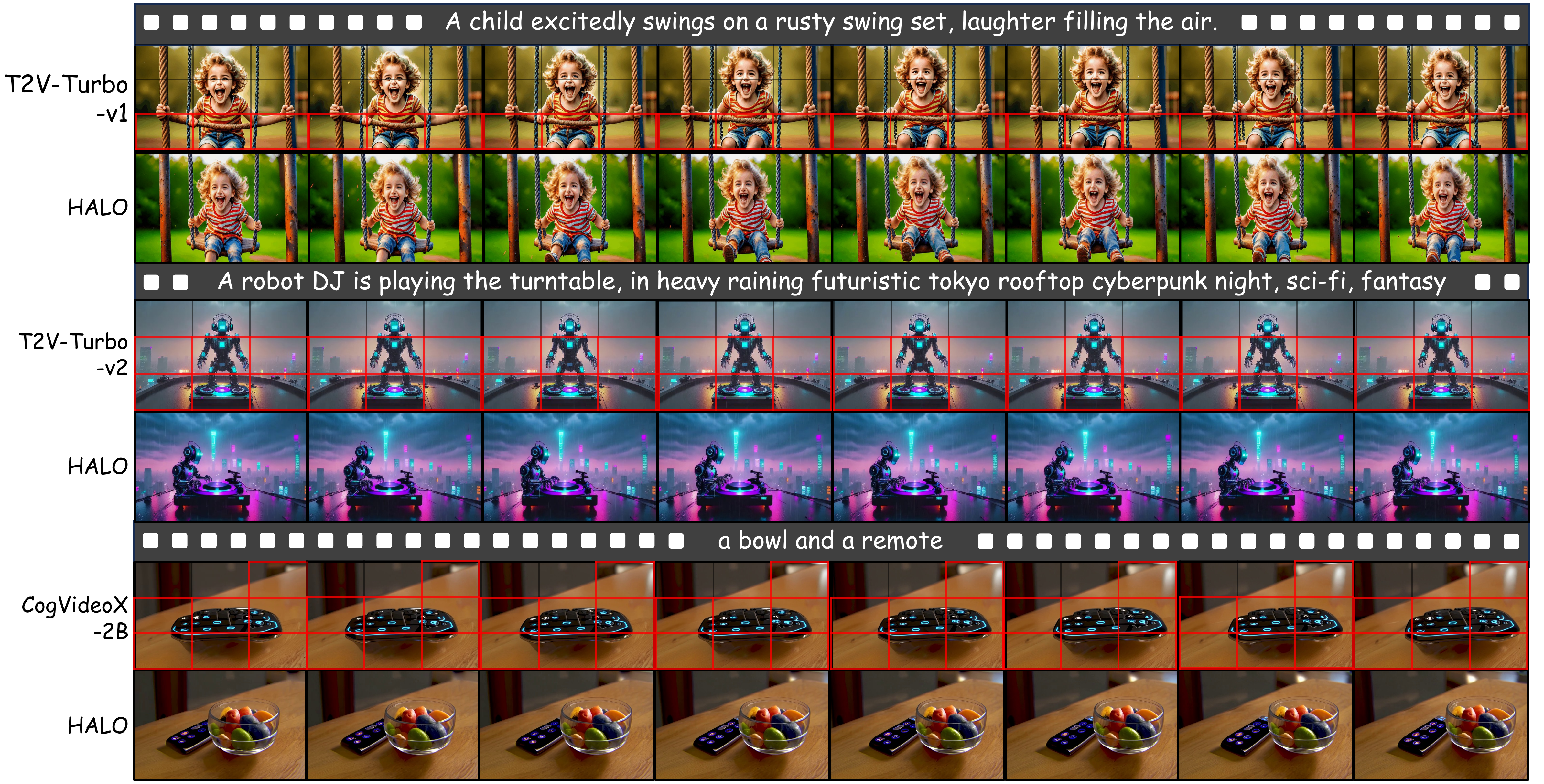}
    \caption{The visualized comparison between our proposed model \ours{} and baselines.
    }
    \label{fig:case}
\end{figure*}

We provide overall experimental results in Table~\ref{tab:overall}, and present case comparisons between the base models and those post-trained by \ours{} in Figure~\ref{fig:case}. 

The quantitative comparison indicates that our proposed method significantly enhances the performance of base VGMs across both evaluation methods. This finding underscores its effectiveness and generalization to various evaluation metrics. 
Notably, our model not only improves the pre-trained VGM (CogVideoX-2B), but also further boosts the performance of VGMs that are post-trained by video rewards, \ie, T2V-Turbo-v1 and v2. This supports our assumption that merely introducing video rewards still overlooks localized patch defects, thereby limiting model capabilities. By introducing our fine-grained patch rewards, we could alleviate this issue and further enhance the overall quality of generated videos.

We indicate the unsatisfying generated video patches by red boxes. The visual comparison distinctly illustrates that though our base models could generate videos from prompts that appear good overall, they occasionally exhibit local errors. For example, in the first video generated by the base model, the child has three hands, which defies natural laws. However, our generated video presents a more reasonable video with a bright and colorful background. In the second case, the robot DJ in the upper video fails to play the turntable, whereas our generated video aligns better with the text prompt. Lastly, our generated video successfully supplements the bowl missed by the base model.

\begin{figure*}[ht]
    \centering
    \includegraphics[width=0.95\linewidth]{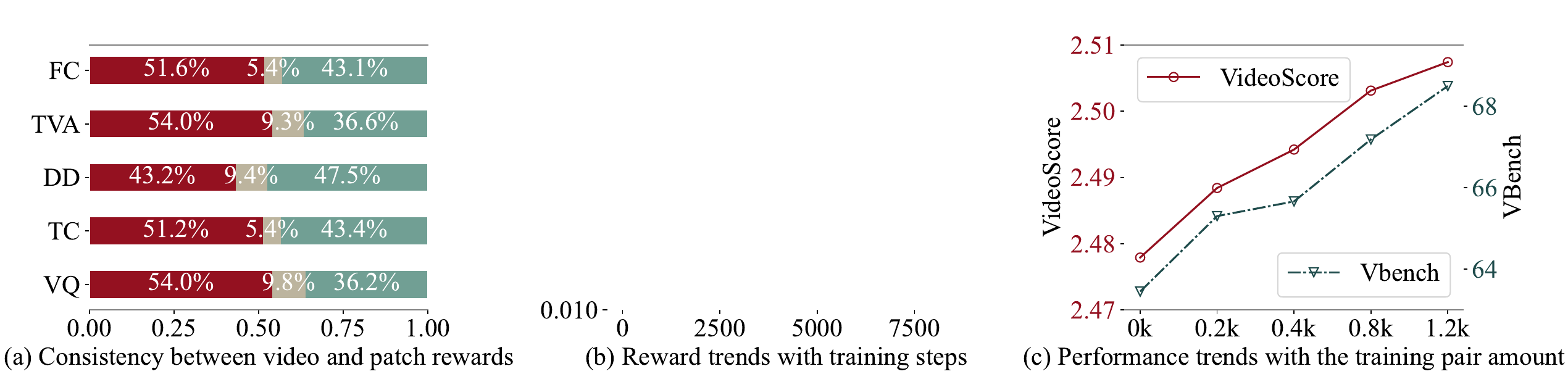}
    \caption{Further analyses about our reward models and training process.
    }
    \label{fig:three-analyses}
\end{figure*}
\subsection{Results of Ablation Experiments}\label{sec:ablation}
We conduct several ablation experiments to validate the impact of our key modules. We examine three model variants by removing components: patch DPO loss (w/o PatchDPO), video DPO loss (w/o VideoDPO), and DPO pair weights (w/o PairWeight). Additionally, we replace our video reward model with two commonly used alternatives, PickScore~\cite{PickScore} and HPSv2~\cite{HPSv2}, resulting in two variants, w/ PickScore and w/ HPSv2. T2V-Turbo-v2 serves as a representative model for experiments due to its fast inference speed. The results are shown in Table~\ref{tab:ablation}. 

The comparison reveals that relying solely on either patch or video DPO results in underperformance compared to our complete model. This phenomenon proves that using local and global rewards collaboratively could complement each other, leading to better generation quality. Meanwhile, the decreased performance of w/o PairWeight compared to \ours{} also reveals that it is necessary to consider the different importance of DPO training pairs. It guides model optimization to focus more on challenging samples, resulting in better generation performance. 
We noticed that changing our video RM also decreases the model performance. This may be attributed to these reward models independently evaluating the video frames, potentially neglecting the temporal evaluation. Additionally, the scales and distributions of these reward models differ from our patch reward model, which may introduce inconsistencies during optimization.  

\subsection{Evaluation of Patch Reward Model}\label{sec:human-eval}
To ensure the effectiveness and reliability of our patch reward model, we perform human annotations to assess the reward quality of GPT-4o and our distilled patch RM. Following VideoScore~\cite{VideoScore}, we define five evaluation dimensions for video patches with scores ranging from 1 to 4. The annotated videos are sampled from VideoScore-collected AI-generated videos and prompts. Finally, valid videos are validly annotated with 360 labeled video patches. The annotation guidelines are presented in Appendix~\ref{app:instructions}. 

Concretely, we calculate Spearman's rank correlation coefficient between two patch reward models to assess their correlations. We demonstrate the evaluation results in Table~\ref{tab:eval_patch_rw}. The evidently positive correlation between the GPT-4o patch RM and human annotations confirms the validity of using GPT-4o as the teacher model to train our patch reward model. Furthermore, our patch reward model exhibits a significantly positive correlation with the GPT-4o patch RM, underscoring the effectiveness of our distillation. Lastly, the positive correlation between the patch reward model and human annotations further validates the reliability of our patch reward model. We also provide cases to show the patch score correlations in Appendix~\ref{app:case-patch-score}.

\begin{figure}[t]
    \centering
    \includegraphics[width=\linewidth]{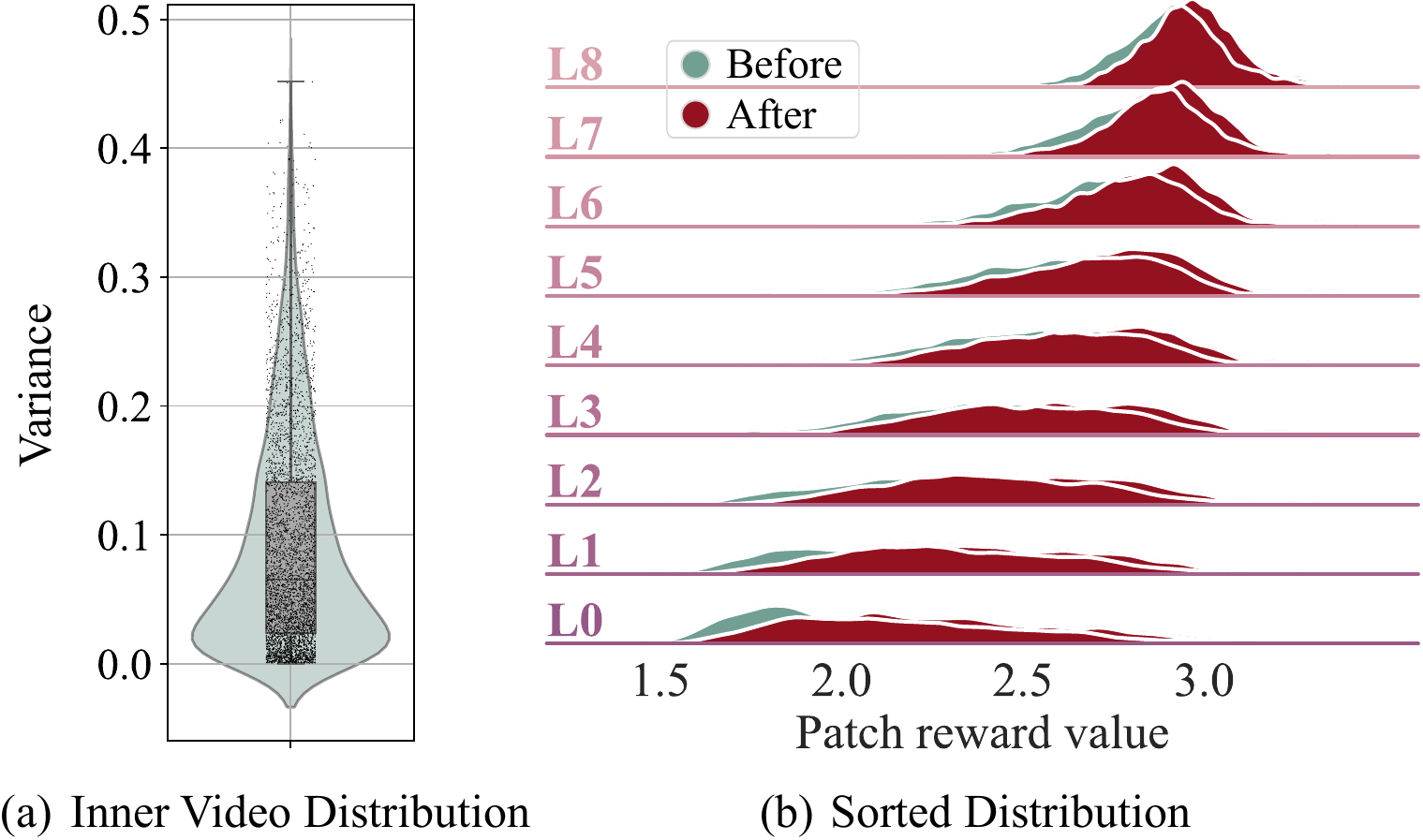}
    \caption{Two types of patch reward distributions. 
    }
    \vspace{-5pt}
    \label{fig:patch-distribution}
\end{figure}
\subsection{Visualization of Reward Distributions}
\paragraph{Consistency distribution of reward preference.}
To quantitatively analyze the importance of considering patch rewards, we conduct visualization experiments on our reward distributions. 
Preliminarily, we compare the preference consistency between video and patch rewards. For a generated video, we obtain its video and patch rewards from our two specialized reward models. We average the patch rewards from nine patches to generate a reward scalar from patch reward models. Then, for five generated videos from the same text prompt, we perform pairwise comparisons of the preference consistency between their video rewards and patch rewards. If the preferences from the video and patch rewards are the same, we label the pair as ``Consistent'', if one of the rewards has no preferences between the two videos (reward values are identical), we label the pair as ``None'', if the preferences differ, they are labeled ``Inverse''. We conduct this analysis on all our training prompts by the results from T2V-Turbo-v2. The distribution of preference consistency is visualized in Figure~\ref{fig:three-analyses} (a). 

From the comparison result, we observe some consistency between video and patch rewards, but they are not exactly the same. This phenomenon further validates the following two key points: (1) The reliability of our patch reward models, which appears to be of a certain correlation with the video rewards. (2) Patch rewards may bring more fine-grained information beyond what is offered by video rewards. Therefore, they do not demonstrate exactly consistent preferences. This phenomenon also reinforces the value of introducing patch rewards to optimize VGMs.

\paragraph{Distribution of patch reward variances.}
Then, we define and visualize two types of patch reward distributions to further analyze our model. The first is the \textit{inner video distribution}, which measures the variance of patch rewards within a video, quantitatively reflecting the patch reward discrepancies inner a video. The second is the \textit{sorted distribution}, which demonstrates the reward distributions from the perspective of sorted patch values. Specifically, for each video with nine patch rewards, we first sort these reward values to obtain nine levels of rewards. ``L0'' means the lowest reward and ``L8'' denotes the highest. Then, we analyze the nine-level rewards across all videos and visualize the distribution of each reward level, leading to nine reward distributions. To reflect the effectiveness of our method, we visualize the sorted distributions both \textit{before} and \textit{after} fine-tuning with  \ours{}. The visualization results of these two distributions are shown in Figure~\ref{fig:patch-distribution}.

From Figure~\ref{fig:patch-distribution}~(a), we observe significant reward variance among different patches within the same video, which confirms that different patches have different generation difficulty and quality. Therefore, it is necessary to consider and model the discrepancies among different patches to provide fine-grained feedback for the optimization of VGMs. According to Figure~\ref{fig:patch-distribution}~(b), 
we notice that distributions of patch values have significant discrepancies among different levels. This phenomenon proves our previous assumption and observation that VGMs usually generate videos containing good overall quality but sometimes have localized errors. After fine-tuning the VGM by our method, the distributions of all levels shift to the right. These results validate that our proposed method not only could improve the overall quality of generated videos but also could enhance the localized and fine-grained quality of AI-generated videos. As a result, it could generate more realistic and reliable videos.

\subsection{Analysis of Training Settings}
We further analyze the detailed performance of the model during training and the impact of different training configurations on the model performance. 
First, we visualize the DPO reward trendlines of our method during the training of our model (the T2V-Turbo-V2 as the representation). Specifically, we demonstrate the reward trendlines of winners and losers in inner DPO pairs. The results are presented in Figure~\ref{fig:three-analyses} (b). 
It is evident that during the training process, the rewards of winners become greater and the loser rewards gradually drop down. This phenomenon proves the effectiveness of our training algorithm and the quality of our constructed DPO pairs that bring valuable information to optimize VGM to focus on fine-grained video generation quality, leading to better model performance.  

Then, we validate the influence of the amount of training data on the model performance. Concretely, we scaled the number of DPO pairs training from 0 to 1.2k with T2V-Turbo-v2 as a representative experiment model. The trends in model performance with the amount of training data are provided in Figure~\ref{fig:three-analyses} (c). Obviously, the average scores of VideoScore and VBench generally increase with increasing amounts of training data. This fits with our intuition because more training data often introduces more valuable information to optimize text-to-video generation models. Due to limited resources and generated training instances, we do not scale our training amount further and leave the expansion experiment to our future work. 
\section{Conclusion}
In this paper, we propose explicitly considering and minimizing localized errors in AI-generated videos, hence developing a post-training framework for text-to-video generation models, \ours{}. 
This framework introduces a patch reward model that delivers fine-grained training signals alongside video rewards to enhance the advanced optimization of VGMs. To efficiently and effectively create the patch reward model, we continuously train our video reward model by distilling GPT-4o-annotated labels. This strategy simplifies the optimization task and yields consistent distributions between patch and video rewards. Furthermore, we develop a Gran-DPO algorithm to post-train diffusion models by collaboratively leveraging patch and video DPO. Our comprehensive experiments demonstrate a positive correlation between our patch reward and human annotations, as well as the superior video generation capabilities of our proposed method compared to baseline models.




\section*{Impact Statement}




This paper focuses primarily on improving the quality of diffusion-model-based text-to-video generation models. Our approach has potential applications across diverse creative fields, including art, design, and entertainment. However, as the capabilities of AI-generated content expand, it is crucial to consider the ethical implications associated with its use. Ensuring the responsible deployment of these technologies is vital to prevent the spread of misinformation and harmful content. To this end, we advocate for the integration of protective measures such as a safety checker module and invisible watermarking, which can help mitigate these risks while promoting the responsible advancement of AI in text-to-video generation. 

\nocite{langley00}

\newpage
\newpage
\bibliography{example_paper}

\begin{thebibliography}{51}
\providecommand{\natexlab}[1]{#1}
\providecommand{\url}[1]{\texttt{#1}}
\expandafter\ifx\csname urlstyle\endcsname\relax
  \providecommand{\doi}[1]{doi: #1}\else
  \providecommand{\doi}{doi: \begingroup \urlstyle{rm}\Url}\fi

\bibitem[Pic(2025)]{Pictory}
The pictory blog, 2025.
\newblock URL \url{https://pictory.ai/blog}.
\newblock Accessed: 2025-01-27.

\bibitem[run(2025)]{runwayml}
Advancing creativity with artificial intelligence., 2025.
\newblock URL \url{https://runwayml.com/}.
\newblock Accessed: 2025-01-27.

\bibitem[syn(2025)]{synthesia}
Ai video editor., 2025.
\newblock URL \url{https://www.synthesia.io/tools/ai-video-editor}.
\newblock Accessed: 2025-01-27.

\bibitem[Bain et~al.(2021)Bain, Nagrani, Varol, and Zisserman]{webvid-1m}
Bain, M., Nagrani, A., Varol, G., and Zisserman, A.
\newblock Frozen in time: {A} joint video and image encoder for end-to-end retrieval.
\newblock In \emph{{ICCV}}, pp.\  1708--1718. {IEEE}, 2021.

\bibitem[Black et~al.(2024)Black, Janner, Du, Kostrikov, and Levine]{DDPO}
Black, K., Janner, M., Du, Y., Kostrikov, I., and Levine, S.
\newblock Training diffusion models with reinforcement learning.
\newblock In \emph{{ICLR}}. OpenReview.net, 2024.

\bibitem[Blattmann et~al.(2023)Blattmann, Rombach, Ling, Dockhorn, Kim, Fidler, and Kreis]{AlignLatent}
Blattmann, A., Rombach, R., Ling, H., Dockhorn, T., Kim, S.~W., Fidler, S., and Kreis, K.
\newblock Align your latents: High-resolution video synthesis with latent diffusion models.
\newblock In \emph{{CVPR}}, pp.\  22563--22575. {IEEE}, 2023.

\bibitem[Chen et~al.(2023)Chen, Xia, He, Zhang, Cun, Yang, Xing, Liu, Chen, Wang, Weng, and Shan]{VC1}
Chen, H., Xia, M., He, Y., Zhang, Y., Cun, X., Yang, S., Xing, J., Liu, Y., Chen, Q., Wang, X., Weng, C., and Shan, Y.
\newblock Videocrafter1: Open diffusion models for high-quality video generation.
\newblock \emph{CoRR}, abs/2310.19512, 2023.

\bibitem[Chen et~al.(2024)Chen, Zhang, Cun, Xia, Wang, Weng, and Shan]{VC2}
Chen, H., Zhang, Y., Cun, X., Xia, M., Wang, X., Weng, C., and Shan, Y.
\newblock Videocrafter2: Overcoming data limitations for high-quality video diffusion models.
\newblock In \emph{{CVPR}}, pp.\  7310--7320. {IEEE}, 2024.

\bibitem[{\c{C}}i{\c{c}}ek et~al.(2016){\c{C}}i{\c{c}}ek, Abdulkadir, Lienkamp, Brox, and Ronneberger]{3DUNet}
{\c{C}}i{\c{c}}ek, {\"{O}}., Abdulkadir, A., Lienkamp, S.~S., Brox, T., and Ronneberger, O.
\newblock 3d u-net: Learning dense volumetric segmentation from sparse annotation.
\newblock In \emph{{MICCAI} {(2)}}, volume 9901 of \emph{Lecture Notes in Computer Science}, pp.\  424--432, 2016.

\bibitem[Clark et~al.(2024)Clark, Vicol, Swersky, and Fleet]{DRaFT}
Clark, K., Vicol, P., Swersky, K., and Fleet, D.~J.
\newblock Directly fine-tuning diffusion models on differentiable rewards.
\newblock In \emph{{ICLR}}. OpenReview.net, 2024.

\bibitem[Fan et~al.(2023)Fan, Watkins, Du, Liu, Ryu, Boutilier, Abbeel, Ghavamzadeh, Lee, and Lee]{DPOK}
Fan, Y., Watkins, O., Du, Y., Liu, H., Ryu, M., Boutilier, C., Abbeel, P., Ghavamzadeh, M., Lee, K., and Lee, K.
\newblock {DPOK:} reinforcement learning for fine-tuning text-to-image diffusion models.
\newblock \emph{CoRR}, abs/2305.16381, 2023.

\bibitem[He et~al.(2024)He, Jiang, Zhang, Ku, Soni, Siu, Chen, Chandra, Jiang, Arulraj, Wang, Do, Ni, Lyu, Narsupalli, Fan, Lyu, Lin, and Chen]{VideoScore}
He, X., Jiang, D., Zhang, G., Ku, M., Soni, A., Siu, S., Chen, H., Chandra, A., Jiang, Z., Arulraj, A., Wang, K., Do, Q.~D., Ni, Y., Lyu, B., Narsupalli, Y., Fan, R., Lyu, Z., Lin, B.~Y., and Chen, W.
\newblock Videoscore: Building automatic metrics to simulate fine-grained human feedback for video generation.
\newblock In \emph{{EMNLP}}, pp.\  2105--2123. Association for Computational Linguistics, 2024.

\bibitem[Ho et~al.(2020)Ho, Jain, and Abbeel]{DM2}
Ho, J., Jain, A., and Abbeel, P.
\newblock Denoising diffusion probabilistic models.
\newblock In \emph{NeurIPS}, 2020.

\bibitem[Ho et~al.(2022)Ho, Salimans, Gritsenko, Chan, Norouzi, and Fleet]{VDM}
Ho, J., Salimans, T., Gritsenko, A., Chan, W., Norouzi, M., and Fleet, D.~J.
\newblock Video diffusion models.
\newblock \emph{arXiv:2204.03458}, 2022.

\bibitem[Hong et~al.(2023)Hong, Ding, Zheng, Liu, and Tang]{CogVideo}
Hong, W., Ding, M., Zheng, W., Liu, X., and Tang, J.
\newblock Cogvideo: Large-scale pretraining for text-to-video generation via transformers.
\newblock In \emph{{ICLR}}. OpenReview.net, 2023.

\bibitem[Hu et~al.(2022)Hu, Shen, Wallis, Allen{-}Zhu, Li, Wang, Wang, and Chen]{lora}
Hu, E.~J., Shen, Y., Wallis, P., Allen{-}Zhu, Z., Li, Y., Wang, S., Wang, L., and Chen, W.
\newblock Lora: Low-rank adaptation of large language models.
\newblock In \emph{{ICLR}}. OpenReview.net, 2022.

\bibitem[Huang et~al.(2024)Huang, He, Yu, Zhang, Si, Jiang, Zhang, Wu, Jin, Chanpaisit, Wang, Chen, Wang, Lin, Qiao, and Liu]{vbench}
Huang, Z., He, Y., Yu, J., Zhang, F., Si, C., Jiang, Y., Zhang, Y., Wu, T., Jin, Q., Chanpaisit, N., Wang, Y., Chen, X., Wang, L., Lin, D., Qiao, Y., and Liu, Z.
\newblock Vbench: Comprehensive benchmark suite for video generative models.
\newblock In \emph{{CVPR}}, pp.\  21807--21818. {IEEE}, 2024.

\bibitem[Jiang et~al.(2024)Jiang, He, Zeng, Wei, Ku, Liu, and Chen]{mantis}
Jiang, D., He, X., Zeng, H., Wei, C., Ku, M., Liu, Q., and Chen, W.
\newblock {MANTIS:} interleaved multi-image instruction tuning.
\newblock \emph{CoRR}, abs/2405.01483, 2024.

\bibitem[Khachatryan et~al.(2023)Khachatryan, Movsisyan, Tadevosyan, Henschel, Wang, Navasardyan, and Shi]{Text2Video-Zero}
Khachatryan, L., Movsisyan, A., Tadevosyan, V., Henschel, R., Wang, Z., Navasardyan, S., and Shi, H.
\newblock Text2video-zero: Text-to-image diffusion models are zero-shot video generators.
\newblock In \emph{{ICCV}}, pp.\  15908--15918. {IEEE}, 2023.

\bibitem[Kirstain et~al.(2023)Kirstain, Polyak, Singer, Matiana, Penna, and Levy]{PickScore}
Kirstain, Y., Polyak, A., Singer, U., Matiana, S., Penna, J., and Levy, O.
\newblock Pick-a-pic: An open dataset of user preferences for text-to-image generation.
\newblock In \emph{NeurIPS}, 2023.

\bibitem[Lai et~al.(2024)Lai, Tian, Chen, Yang, Peng, and Jia]{Step-DPO}
Lai, X., Tian, Z., Chen, Y., Yang, S., Peng, X., and Jia, J.
\newblock Step-dpo: Step-wise preference optimization for long-chain reasoning of llms.
\newblock \emph{CoRR}, abs/2406.18629, 2024.

\bibitem[Lee et~al.(2023)Lee, Liu, Ryu, Watkins, Du, Boutilier, Abbeel, Ghavamzadeh, and Gu]{AlignDM}
Lee, K., Liu, H., Ryu, M., Watkins, O., Du, Y., Boutilier, C., Abbeel, P., Ghavamzadeh, M., and Gu, S.~S.
\newblock Aligning text-to-image models using human feedback.
\newblock \emph{CoRR}, abs/2302.12192, 2023.

\bibitem[Li et~al.(2024{\natexlab{a}})Li, Feng, Fu, Wang, Basu, Chen, and Wang]{t2v-turbo}
Li, J., Feng, W., Fu, T., Wang, X., Basu, S., Chen, W., and Wang, W.~Y.
\newblock T2v-turbo: Breaking the quality bottleneck of video consistency model with mixed reward feedback.
\newblock \emph{CoRR}, abs/2405.18750, 2024{\natexlab{a}}.

\bibitem[Li et~al.(2024{\natexlab{b}})Li, Long, Zheng, Gao, Piramuthu, Chen, and Wang]{t2v-turbo-v2}
Li, J., Long, Q., Zheng, J., Gao, X., Piramuthu, R., Chen, W., and Wang, W.~Y.
\newblock T2v-turbo-v2: Enhancing video generation model post-training through data, reward, and conditional guidance design.
\newblock \emph{CoRR}, abs/2410.05677, 2024{\natexlab{b}}.

\bibitem[Li et~al.(2023)Li, Yang, and Wang]{rlhf}
Li, Z., Yang, Z., and Wang, M.
\newblock Reinforcement learning with human feedback: Learning dynamic choices via pessimism.
\newblock \emph{CoRR}, abs/2305.18438, 2023.

\bibitem[Miao et~al.(2024)Miao, Wang, Wang, Yang, Wang, Qiu, and Liu]{RLDMDiverse}
Miao, Z., Wang, J., Wang, Z., Yang, Z., Wang, L., Qiu, Q., and Liu, Z.
\newblock Training diffusion models towards diverse image generation with reinforcement learning.
\newblock In \emph{{CVPR}}, pp.\  10844--10853. {IEEE}, 2024.

\bibitem[Nan et~al.(2024)Nan, Xie, Zhou, Fan, Yang, Chen, Li, Yang, and Tai]{OpenVid-1M}
Nan, K., Xie, R., Zhou, P., Fan, T., Yang, Z., Chen, Z., Li, X., Yang, J., and Tai, Y.
\newblock Openvid-1m: {A} large-scale high-quality dataset for text-to-video generation.
\newblock \emph{CoRR}, abs/2407.02371, 2024.

\bibitem[OpenAI(2023)]{GPT-4}
OpenAI.
\newblock {GPT-4} technical report.
\newblock \emph{CoRR}, abs/2303.08774, 2023.

\bibitem[Ouyang et~al.(2022{\natexlab{a}})Ouyang, Wu, Jiang, Almeida, Wainwright, Mishkin, Zhang, Agarwal, Slama, Ray, Schulman, Hilton, Kelton, Miller, Simens, Askell, Welinder, Christiano, Leike, and Lowe]{GPT-3.5}
Ouyang, L., Wu, J., Jiang, X., Almeida, D., Wainwright, C.~L., Mishkin, P., Zhang, C., Agarwal, S., Slama, K., Ray, A., Schulman, J., Hilton, J., Kelton, F., Miller, L., Simens, M., Askell, A., Welinder, P., Christiano, P.~F., Leike, J., and Lowe, R.
\newblock Training language models to follow instructions with human feedback.
\newblock In \emph{NeurIPS}, 2022{\natexlab{a}}.

\bibitem[Ouyang et~al.(2022{\natexlab{b}})Ouyang, Wu, Jiang, Almeida, Wainwright, Mishkin, Zhang, Agarwal, Slama, Ray, Schulman, Hilton, Kelton, Miller, Simens, Askell, Welinder, Christiano, Leike, and Lowe]{LMFI}
Ouyang, L., Wu, J., Jiang, X., Almeida, D., Wainwright, C.~L., Mishkin, P., Zhang, C., Agarwal, S., Slama, K., Ray, A., Schulman, J., Hilton, J., Kelton, F., Miller, L., Simens, M., Askell, A., Welinder, P., Christiano, P.~F., Leike, J., and Lowe, R.
\newblock Training language models to follow instructions with human feedback.
\newblock In \emph{NeurIPS}, 2022{\natexlab{b}}.

\bibitem[Peebles \& Xie(2023)Peebles and Xie]{DiT}
Peebles, W. and Xie, S.
\newblock Scalable diffusion models with transformers.
\newblock In \emph{{ICCV}}, pp.\  4172--4182. {IEEE}, 2023.

\bibitem[Podell et~al.(2024)Podell, English, Lacey, Blattmann, Dockhorn, M{\"{u}}ller, Penna, and Rombach]{SDXL}
Podell, D., English, Z., Lacey, K., Blattmann, A., Dockhorn, T., M{\"{u}}ller, J., Penna, J., and Rombach, R.
\newblock {SDXL:} improving latent diffusion models for high-resolution image synthesis.
\newblock In \emph{{ICLR}}. OpenReview.net, 2024.

\bibitem[Rombach et~al.(2022)Rombach, Blattmann, Lorenz, Esser, and Ommer]{LatentDM}
Rombach, R., Blattmann, A., Lorenz, D., Esser, P., and Ommer, B.
\newblock High-resolution image synthesis with latent diffusion models.
\newblock In \emph{{CVPR}}, pp.\  10674--10685. {IEEE}, 2022.

\bibitem[Ronneberger et~al.(2015)Ronneberger, Fischer, and Brox]{UNet}
Ronneberger, O., Fischer, P., and Brox, T.
\newblock U-net: Convolutional networks for biomedical image segmentation.
\newblock In \emph{{MICCAI} {(3)}}, volume 9351 of \emph{Lecture Notes in Computer Science}, pp.\  234--241. Springer, 2015.

\bibitem[Schuhmann et~al.(2022)Schuhmann, Beaumont, Vencu, Gordon, Wightman, Cherti, Coombes, Katta, Mullis, Wortsman, Schramowski, Kundurthy, Crowson, Schmidt, Kaczmarczyk, and Jitsev]{LAION-5B}
Schuhmann, C., Beaumont, R., Vencu, R., Gordon, C., Wightman, R., Cherti, M., Coombes, T., Katta, A., Mullis, C., Wortsman, M., Schramowski, P., Kundurthy, S., Crowson, K., Schmidt, L., Kaczmarczyk, R., and Jitsev, J.
\newblock {LAION-5B:} an open large-scale dataset for training next generation image-text models.
\newblock In \emph{NeurIPS}, 2022.

\bibitem[Sohl{-}Dickstein et~al.(2015)Sohl{-}Dickstein, Weiss, Maheswaranathan, and Ganguli]{DM1}
Sohl{-}Dickstein, J., Weiss, E.~A., Maheswaranathan, N., and Ganguli, S.
\newblock Deep unsupervised learning using nonequilibrium thermodynamics.
\newblock In \emph{{ICML}}, volume~37 of \emph{{JMLR} Workshop and Conference Proceedings}, pp.\  2256--2265. JMLR.org, 2015.

\bibitem[Song et~al.(2021)Song, Meng, and Ermon]{DDIM}
Song, J., Meng, C., and Ermon, S.
\newblock Denoising diffusion implicit models.
\newblock In \emph{{ICLR}}. OpenReview.net, 2021.

\bibitem[Song et~al.(2023)Song, Dhariwal, Chen, and Sutskever]{CM}
Song, Y., Dhariwal, P., Chen, M., and Sutskever, I.
\newblock Consistency models.
\newblock In \emph{{ICML}}, volume 202 of \emph{Proceedings of Machine Learning Research}, pp.\  32211--32252. {PMLR}, 2023.

\bibitem[Soomro et~al.(2012)Soomro, Zamir, and Shah]{UCF101}
Soomro, K., Zamir, A.~R., and Shah, M.
\newblock {UCF101:} {A} dataset of 101 human actions classes from videos in the wild.
\newblock \emph{CoRR}, abs/1212.0402, 2012.

\bibitem[Wallace et~al.(2024{\natexlab{a}})Wallace, Dang, Rafailov, Zhou, Lou, Purushwalkam, Ermon, Xiong, Joty, and Naik]{DMDPO}
Wallace, B., Dang, M., Rafailov, R., Zhou, L., Lou, A., Purushwalkam, S., Ermon, S., Xiong, C., Joty, S., and Naik, N.
\newblock Diffusion model alignment using direct preference optimization.
\newblock In \emph{{CVPR}}, pp.\  8228--8238. {IEEE}, 2024{\natexlab{a}}.

\bibitem[Wallace et~al.(2024{\natexlab{b}})Wallace, Dang, Rafailov, Zhou, Lou, Purushwalkam, Ermon, Xiong, Joty, and Naik]{DPODM}
Wallace, B., Dang, M., Rafailov, R., Zhou, L., Lou, A., Purushwalkam, S., Ermon, S., Xiong, C., Joty, S., and Naik, N.
\newblock Diffusion model alignment using direct preference optimization.
\newblock In \emph{{CVPR}}, pp.\  8228--8238. {IEEE}, 2024{\natexlab{b}}.

\bibitem[Wang et~al.(2023{\natexlab{a}})Wang, Yuan, Chen, Zhang, Wang, and Zhang]{ModelScope}
Wang, J., Yuan, H., Chen, D., Zhang, Y., Wang, X., and Zhang, S.
\newblock Modelscope text-to-video technical report.
\newblock \emph{CoRR}, abs/2308.06571, 2023{\natexlab{a}}.

\bibitem[Wang et~al.(2024)Wang, Yu, Wang, Chen, Zhu, and Dou]{richrag}
Wang, S., Yu, X., Wang, M., Chen, W., Zhu, Y., and Dou, Z.
\newblock Richrag: Crafting rich responses for multi-faceted queries in retrieval-augmented generation.
\newblock \emph{CoRR}, abs/2406.12566, 2024.

\bibitem[Wang et~al.(2023{\natexlab{b}})Wang, Chen, Ma, Zhou, Huang, Wang, Yang, He, Yu, Yang, Guo, Wu, Si, Jiang, Chen, Loy, Dai, Lin, Qiao, and Liu]{LAVIE}
Wang, Y., Chen, X., Ma, X., Zhou, S., Huang, Z., Wang, Y., Yang, C., He, Y., Yu, J., Yang, P., Guo, Y., Wu, T., Si, C., Jiang, Y., Chen, C., Loy, C.~C., Dai, B., Lin, D., Qiao, Y., and Liu, Z.
\newblock {LAVIE:} high-quality video generation with cascaded latent diffusion models.
\newblock \emph{CoRR}, abs/2309.15103, 2023{\natexlab{b}}.

\bibitem[Wang et~al.(2023{\natexlab{c}})Wang, Kordi, Mishra, Liu, Smith, Khashabi, and Hajishirzi]{self-instruct}
Wang, Y., Kordi, Y., Mishra, S., Liu, A., Smith, N.~A., Khashabi, D., and Hajishirzi, H.
\newblock Self-instruct: Aligning language models with self-generated instructions.
\newblock In \emph{{ACL} {(1)}}, pp.\  13484--13508. Association for Computational Linguistics, 2023{\natexlab{c}}.

\bibitem[Wu et~al.(2023)Wu, Hao, Sun, Chen, Zhu, Zhao, and Li]{HPSv2}
Wu, X., Hao, Y., Sun, K., Chen, Y., Zhu, F., Zhao, R., and Li, H.
\newblock Human preference score v2: {A} solid benchmark for evaluating human preferences of text-to-image synthesis.
\newblock \emph{CoRR}, abs/2306.09341, 2023.

\bibitem[Xu et~al.(2024)Xu, Zou, Huang, Chen, Liu, Cheng, Shi, and Huang]{EasyAnimate-DiT}
Xu, J., Zou, X., Huang, K., Chen, Y., Liu, B., Cheng, M., Shi, X., and Huang, J.
\newblock Easyanimate: {A} high-performance long video generation method based on transformer architecture.
\newblock \emph{CoRR}, abs/2405.18991, 2024.

\bibitem[Yang et~al.(2024)Yang, Teng, Zheng, Ding, Huang, Xu, Yang, Hong, Zhang, Feng, Yin, Gu, Zhang, Wang, Cheng, Liu, Xu, Dong, and Tang]{CogVideoX}
Yang, Z., Teng, J., Zheng, W., Ding, M., Huang, S., Xu, J., Yang, Y., Hong, W., Zhang, X., Feng, G., Yin, D., Gu, X., Zhang, Y., Wang, W., Cheng, Y., Liu, T., Xu, B., Dong, Y., and Tang, J.
\newblock Cogvideox: Text-to-video diffusion models with an expert transformer.
\newblock \emph{CoRR}, abs/2408.06072, 2024.

\bibitem[Yoon et~al.(2024)Yoon, Yoon, Eom, Han, Nam, Jo, On, Hasegawa{-}Johnson, Kim, and Yoo]{TLCR}
Yoon, E., Yoon, H.~S., Eom, S., Han, G., Nam, D.~W., Jo, D., On, K., Hasegawa{-}Johnson, M., Kim, S., and Yoo, C.~D.
\newblock {TLCR:} token-level continuous reward for fine-grained reinforcement learning from human feedback.
\newblock In \emph{{ACL} (Findings)}, pp.\  14969--14981. Association for Computational Linguistics, 2024.

\bibitem[Yuan et~al.(2024)Yuan, Zhang, Wang, Wei, Feng, Pan, Zhang, Liu, Albanie, and Ni]{InstructVideo}
Yuan, H., Zhang, S., Wang, X., Wei, Y., Feng, T., Pan, Y., Zhang, Y., Liu, Z., Albanie, S., and Ni, D.
\newblock Instructvideo: Instructing video diffusion models with human feedback.
\newblock In \emph{{CVPR}}, pp.\  6463--6474. {IEEE}, 2024.

\bibitem[Zhang et~al.(2023)Zhang, Wu, Liu, Zhao, Ran, Gu, Gao, and Shou]{show-1}
Zhang, D.~J., Wu, J.~Z., Liu, J., Zhao, R., Ran, L., Gu, Y., Gao, D., and Shou, M.~Z.
\newblock Show-1: Marrying pixel and latent diffusion models for text-to-video generation.
\newblock \emph{CoRR}, abs/2309.15818, 2023.

\end{thebibliography}
\bibliographystyle{icml2025}

\newpage
\appendix
\onecolumn
\section{Implementation Details}\label{app:imp_detail}  All our models are trained on 4 NVIDIA A100 GPUs. For the training instance generation, the filtering threshold, $tau$, is set as $0.85$. 878 text prompts were generated by prompting GPT-3.5-Turbo. We split 50 prompts from generated prompts to build the valid set. Then, we inferred our base models to sample five videos for each text prompt to build training text-video instances. 
For the training of the video patch construction, we set $h_n=3$ and $w_n=3$. For the patch reward construction, since we aim to post-train the VideoScore-v1.1, which already has strong video evaluation abilities, we prompt GPT-4o to evaluate video patches from 492 text-video pairs (sampled from training data) to decrease API cost. It contains 4,428 labeled video patches. 
We split them by the ratio 6:1:3 into training, validation, and test datasets to continually train VideoScore-v1.1, building our patch reward model. The batch size is 64, the learning rate is 1e-6, and the epoch is 10. 
Then we evaluate all generated text-video instances by our reward models to build training pairs for Gran-DPO. To ensure effective and efficient post-training, we used the LoRA technique~\cite{lora} to conduct the post-training. Specifically, the LoRA rank was set by 64, the learning rate was set by 1e-4, and the training step was 8k. Following the setting of base models, the inference diffusion steps of T2V-Turbo-v1 and v2 are 8, and CogVideoX-2B is 50. 

\section{Case Study of Patch Scores}\label{app:case-patch-score}
\begin{figure*}
    \centering
    \includegraphics[width=1\linewidth]{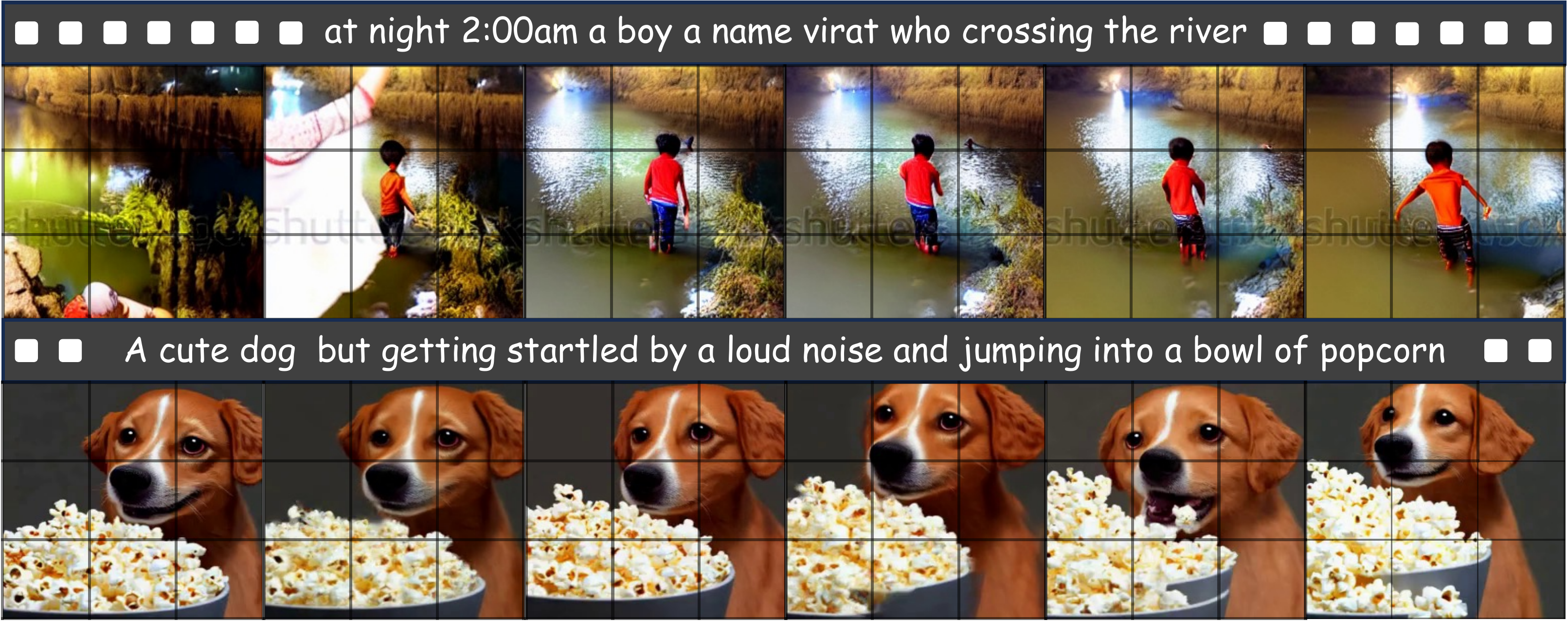}
    \caption{Cases to show the evaluated patch rewards.}
    \label{fig:app-case}
\end{figure*}
\begin{table}[th]
    \centering
    \caption{The cases to show patch rewards from human annotations and GPT-4o. ``0'', ``1'', and ``2'' denote the indexes of patch columns and rows.}
    \begin{tabular}{>{\centering\arraybackslash}p{0.07\textwidth}|>{\centering\arraybackslash}p{0.12\textwidth}>
    {\centering\arraybackslash}p{0.12\textwidth}>
    {\centering\arraybackslash}p{0.12\textwidth}|>
    {\centering\arraybackslash}p{0.1\textwidth}>
    {\centering\arraybackslash}p{0.1\textwidth}>
    {\centering\arraybackslash}p{0.1\textwidth}}
        \toprule
        - & \multicolumn{3}{c|}{\makecell{at night 2:00am a boy a name virat who crossing\\ the river and also he enjoys the changing seasons \\amidst the slow waves of the river.}} & \multicolumn{3}{c}{\makecell{A cute dog  but getting startled by a loud \\noise and jumping into a bowl of popcorn}} \\
        \midrule
        - & 0 & 1 & 2 & 0 & 1 & 2  \\
        \midrule
        0 & 1.8/1.2 & 2.2/1.2 & 2.2/1.6 & 1.8/1.0 & 1.8/2.3 & 1.8/2.3 \\
        \midrule
        1 & 1.8/1.0 & 1.8/1.4 & 2.2/1.2 & 1.6/1.9 & 1.8/2.2 & 1.6/1.4 \\
        \midrule
        2 & 1.8/1.1 & 1.8/1.4 & 1.8/1.0 & 1.6/1.7 & 1.6/1.5 & 1.6/1.5 \\
        \bottomrule
    \end{tabular}
    \label{tab:app-case}
\end{table}

We also demonstrate some cases to show the positive correlations between GPT-4o patch scores and human annotations. For the shown AI-generated videos, which are collected by VideoScore, we present the video contents in Figure~\ref{fig:app-case} and the patch reward scores in Table~\ref{tab:app-case}. Note that the complete prompt of the first case is ``at night 2:00am a boy a name virat who crossing the river and also he enjoys the changing seasons amidst the slow waves of the river.''. 

In the first case, there is a sudden ``white back'' in the left regions. Consequently, the patch scores of these regions, such as ``(0,0)'',``(1,0)'',``(2,0)'', and ``(2,1)'' are relatively low in both human annotation and GPT-4o scores. In the second case, some patches of the ``double-walled bowl'' \eg, ``(2,1)'' and ``(2,2)'', are in contrast to the real world, thus their patch scores are relatively low in both reward methods. The results also demonstrate a certain positive correlation between GPT-4o and human annotations, confirming its reliability in serving as our teacher model.

\section{Details of Used Instructions}\label{app:instructions}
\begin{figure*}[h]
    \begin{tcolorbox}[title={Instructions for Human Annotators to Label Video Patch Scores.}] 
- Task Brief: judging and evaluating the quality of AI-generated videos (generated by providing a text prompt to AI models).\newline
- Description: \newline
    - For an AI-generated video, we divide 9 video patches along the height and width of the video according to a 3*3 grid. I will provide you with the **text prompt**, the **generated video**, and its **video patches** with coordinates (the coordinate of the left-upper patch is (0,0)). \newline
    - Please give patch-level fine-grained scores for all video patches to represent the generation quality of this video patch. \newline
    - Please consider the following 5 dimensions to evaluate the video patch:\newline
        (1) visual quality: the quality of the video patch in terms of clearness, resolution, brightness, and color.\newline
        (2) temporal consistency: the consistency of objects or humans or background in video patch.\newline
        (3) dynamic degree: the degree of dynamic changes.\newline
        (4) text-to-video alignment: the alignment between the text prompt and the video patch content. Note that please only evaluate the semantic consistency between the **video patch** and **the text part related to this patch**, instead of the whole text (The semantic consistency between a video patch and the whole input text should always be very low).\newline
        (5) factual consistency: the consistency of the video patch content with the common-sense and factual knowledge.\newline
    - Annotation Format: For each dimension, output a number from [1,2,3,4],
        in which '1' means 'Bad', '2' means 'Average', '3' means 'Good', 
        '4' means 'Real' or 'Perfect' (the video is like a real video)\newline
    - Note: \newline
        1. I will provide the evaluation scores of these 5 dimensions of the original video, which can provide some reference for the evaluation of the video patch.\newline
        2. Please write the evaluation result in the file named "evaluation\_results.xlsx".\newline
    \end{tcolorbox}
\caption{Instructions for human annotators to label video patch scores.}
\label{box:human-annotation-instruction}
\end{figure*}

\begin{figure*}[h]
    \begin{tcolorbox}[title={Instruction for GPT-3.5-Turbo to Generate Training Text Prompts}] 
    \#\# Background\newline
    You are an expert in generating user queries. I will provide you with an evaluation dimension, and your task is to generate user queries based on this dimension. The generated queries will be used as input for video generation models, allowing me to evaluate the quality of the generated videos based on the specified dimension. Your goal is to generate high-quality and diverse text queries that align with the provided evaluation dimension.\newline
    
    \#\#  Input Format\newline
    I will provide you with an evaluation dimension and some good examples in the following JSON format:\newline
    \{\{\newline
    \quad"dimension": a string indicating the evaluation dimension,\newline
    \quad"examples": a list of JSON-formatted data pieces, each representing a good example for the given dimension\newline
    \}\}\newline
    \newline
    \#\#  Output Format\newline
    You should generate a list of high-quality and diverse query examples that are closely related to the provided dimension and examples. The generated examples should follow the writing style and tone of the provided examples. The output should be in the format of a JSON list, with the same data format as the examples in the input.\newline
    \newline
    \#\#\# Notions\newline
    1. Closely follow the writing style and tone of the provided examples when generating new queries.\newline
    2. Output only the JSON-formatted list of generated queries, without any additional words or characters.\newline
    \end{tcolorbox}
\caption{Instruction for GPT-4o to Generate Evaluation Scores for the Input Video Patch.}
\label{box:text-generation-instruction}
\end{figure*}

\begin{figure*}[h]
    \begin{tcolorbox}[title={Instruction of GPT-4o to generate evaluation scores for the input video patch.}] 
    \#\# Background \newline
    Suppose you are an expert in judging and evaluating the quality of AI-generated videos. For a generated video, we divide 9 video patches along the height and width of the video according to a 3*3 grid. I will provide you with the text prompt, the generated video, one of its video patches, and the coordinate of the video patch (the coordinate of the left-upper patch is (0,0), the first number means the index of rows, and the second number means the index of columns). Please watch the following frames of a given video and one of its split patches, and see the text prompt for generating the video. Then, you should give patch-level fine-grained scores for this video patch to represent the generation quality of this video patch. The evaluation of the video patch\'s quality should cover the following dimensions: \newline
        \{\{ \newline
            "visual quality": "the quality of the video in terms of clearness, resolution, brightness, and color.", \newline
            "temporal consistency": "the consistency of objects or humans in video.", \newline
            "dynamic degree": "the degree of dynamic changes.", \newline
            "text-to-video alignment": "the alignment between the text prompt and the video content.", \newline
            "factual consistency": "the consistency of the video content with the common-sense and factual knowledge." \newline
        \}\}
     \newline
    \#\#  Evaluation Requirments \newline
    1. You should do **fine-grained and patch-leval** evaluation for **assessing the quality of the video patch, instead of the whole video**. The role of the entire video is to provide the context of the video patch to understand it better.   \newline
    2. You should detail view the patch of the input video to give a reliable and rigorous evaluation score of this patch, both considering the patch\'s local quality and its correlation with the whole video. \newline
    3. You should carefully evaluate the reality of the generated video patch to provide reliable evaluation results. \newline
     \newline
    \#\#  Output Requirments \newline
    1. For each dimension, output a number ranging from 0 to 10, in which ’0’ means the worst and '10' means the best.  \newline
    2. The output should be in JSON format. \newline
    3. Only output the JSON-format result without any other words. \newline
    4. Here is an output example: \newline
        \{\{ \newline
            "visual quality": An int value ranging from 0 to 10, representing the visual quality score of the evaluated video, \newline
            "temporal consistency": An int value ranging from 0 to 10, representing the temporal consistency score of the evaluated video, \newline
            "dynamic degree": An int value ranging from 0 to 10, representing the dynamic degree score of the evaluated video, \newline
            "text-to-video alignment": An int value ranging from 0 to 10, representing the text-to-video alignment score of the evaluated video, \newline
            "factual consistency": An int value ranging from 0 to 10, representing the factual consistency score of the evaluated video \newline
        \}\}
     \newline
    \#\#  Input information \newline
    - Text prompt: "\{text\_prompt\}". \newline
    - Coordinate of the patch: (\{m\}, \{n\}). \newline
    - All the frames of the video and its evaluated patch are as follows: \newline
    \end{tcolorbox}
\caption{Instruction for GPT-4o to generate evaluation scores for the input video patch.}
\label{box:gpt4-patch-eval}
\end{figure*}




\end{document}